\pdfoutput=1
\documentclass{article}
\usepackage[utf8]{inputenc}
\usepackage{config/spconf,amsmath,amssymb,graphicx,hyperref,multirow,array,booktabs,xcolor,colortbl,wrapfig}

\usepackage{enumitem}
\setlist{nosep, leftmargin=*, topsep=1pt, partopsep=0pt}

\hypersetup{hidelinks}
\renewcommand{\refname}{REFERENCES}



\title{Phys-Diff: A Physics-Inspired Latent Diffusion Model for Tropical Cyclone Forecasting}
\name{Lei Liu$^1$, Xiaoning Yu$^2$, Kang Chen$^1$, Jiahui Huang$^1$, Tengyuan Liu$^1$, Hongwei Zhao$^1$\sthanks{Corresponding author. The source code is available at: \url{https://github.com/USTC-AI4EEE/Phys-Diff}.}, Bin Li$^{1}$}
\address{$^1$University of Science and Technology of China, Hefei, China\\
         $^2$Hefei University of Technology, Hefei, China\\
         \{liulei13,hwzhao,binli\}@ustc.edu.cn, xiaoningyu3@gmail.com,\\
         \{ck6,TengyuanLiu\}@mail.ustc.edu.cn, jiahuihuang@mail.ustc.edu.cn}
\begin{document}
\ninept % Comment out this line to use the default 10pt font.
\maketitle
\begin{abstract}
Tropical cyclone (TC) forecasting is critical for disaster warning and emergency response. Deep learning methods address computational challenges but often neglect physical relationships between TC attributes, resulting in predictions lacking physical consistency. To address this, we propose Phys-Diff, a physics-inspired latent diffusion model that disentangles latent features into task-specific components (trajectory, pressure, wind speed) and employs cross-task attention to introduce prior physics-inspired inductive biases, thereby embedding physically consistent dependencies among TC attributes. Phys-Diff integrates multimodal data including historical cyclone attributes, ERA5 reanalysis data, and FengWu forecast fields via a Transformer encoder-decoder architecture, further enhancing forecasting performance. Experiments demonstrate state-of-the-art performance on global and regional datasets.
\end{abstract}
\vspace{-0.4cm}
\section{Introduction}
\vspace{-0.35cm}
\label{sec:intro}

Tropical Cyclones (TCs) are complex weather systems that cause significant damage through strong winds, heavy rainfall, and flooding. Accurate forecasting of their attributes, typically trajectory (latitude and longitude), central pressure, and maximum sustained wind speed, is crucial for disaster prevention and early warning. Current TC forecasting approaches fall into two categories: Numerical Weather Prediction (NWP) models and Deep Learning (DL) methods. NWP models simulate atmospheric dynamics using fundamental physical equations \cite{coiffier2011fundamentals, hakim2024dynamical}. While grounded in physics, they are computationally expensive, often requiring supercomputers, which limits their efficiency for rapid, high-resolution forecasts \cite{bauer2015quiet}. Moreover, due to necessary parameterizations and simplifications, they struggle to capture the complex, non-linear physical relationships among TC attributes.

\begin{figure}[!t]
\centering
\includegraphics[width=0.28\textwidth]{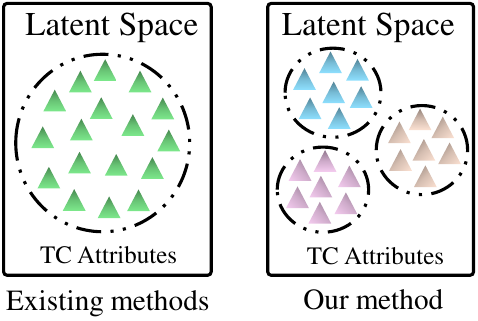}
\caption{Comparison of feature representations. Previous methods (e.g., MSCAR \cite{wang2024mscar}, VQLTI \cite{Wang_Liu_Chen_Han_Li_Bai_2025}) produce coupled features without physical constraints. Our method enforces these constraints, learning disentangled features with improved physical consistency.}
\label{fig:import}
\end{figure}

As an alternative, DL methods like Recurrent Neural Networks (RNNs) and Transformers have shown great potential in capturing non-linear patterns from data at a lower computational cost \cite{gao2018nowcasting, kim2019deep, liu2024hybrid, meng2023probabilistic}. However, many existing DL methods face a critical challenge: they often treat TC attributes independently, ignoring the physical constraints and interdependencies between them \cite{hu2024improving}. This leads to predictions that lack physical consistency and suffer from significant error accumulation in long-term forecasts. As illustrated in Fig.~\ref{fig:import}, this oversight results in an entangled latent space where the distinct features of interdependent attributes are conflated, rather than being clearly represented and systematically related.

Recently, Denoising Diffusion Probabilistic Models (DDPMs) have begun to attract attention in meteorological forecasting for their ability to model complex data distributions and handle the inherent uncertainty in atmospheric systems \cite{ren2025improvingtropicalcycloneforecasting, li2024generative}. However, their application to TC forecasting faces a critical challenge: how to effectively embed physical constraints directly into the generative (denoising) process. Simply conditioning the model on historical data is insufficient to promote physical laws on the generated future states.

To address these issues, we introduce Phys-Diff, a physics-inspired Latent Diffusion Model. At its core, Phys-Diff incorporates physical constraints through its Physics-Inspired Gated Attention (PIGA) module. This module explicitly models the physical interdependencies by first learning disentangled features for each TC attribute and then using a cross-task attention mechanism to simulate their interactions at the feature level, ensuring physical consistency is maintained throughout the prediction process.
This paper makes the following contributions:

\begin{itemize}
\item We introduce Phys-Diff, the first physics-inspired diffusion framework for the joint prediction of TC trajectory and intensity. This approach pioneers embedding physical constraints within the diffusion generative process, paired with an adaptive multi-task loss balancing mechanism to ensure stable training and improve overall prediction accuracy.

\item We design the PIGA module, which explicitly models the physical interactions between multiple TC attributes in the latent space, improving feature representation quality and ensuring physical consistency in predictions.

\item Phys-Diff achieves state-of-the-art performance, reducing the 24-hour forecast error on the global dataset for trajectory by 41.6\%, pressure by 57.1\%, and wind speed by 71.2\% against the best competing deep learning models.
\end{itemize}

\begin{figure*}[t]
\centering
\includegraphics[width=1\textwidth]{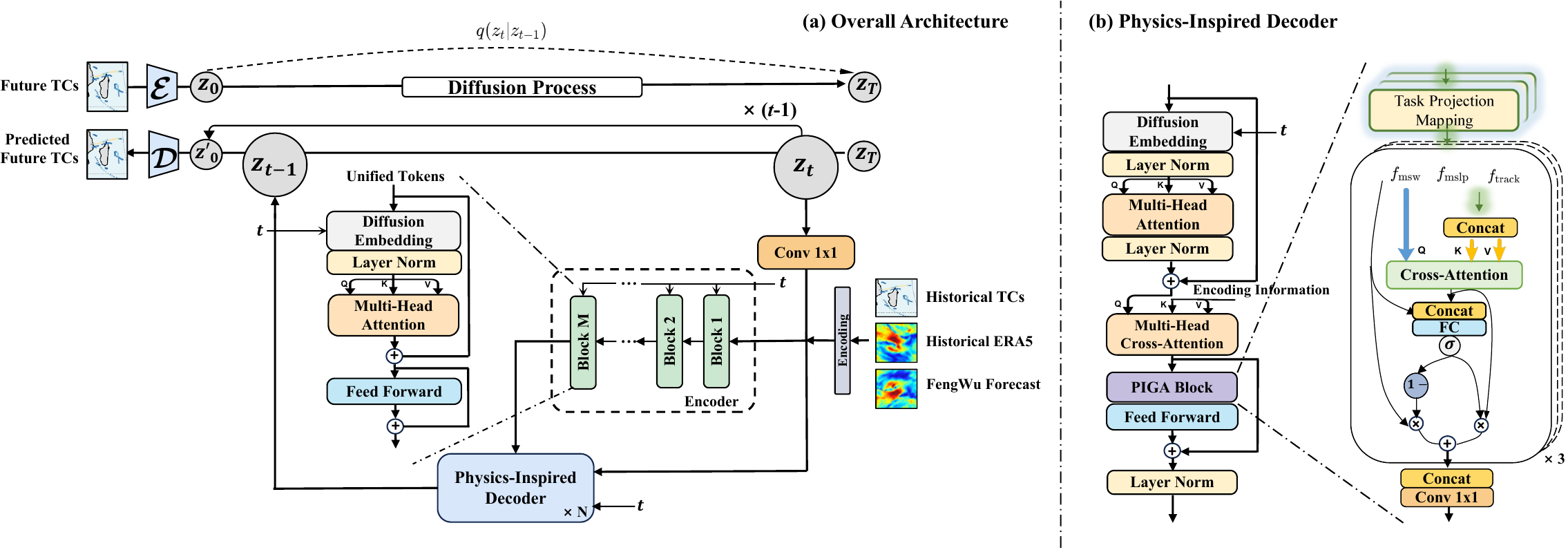}
\vspace{-22pt}
\caption{Overall architecture of the Phys-Diff model. This is a denoising diffusion model built on a Transformer encoder-decoder framework. During inference, the model starts with random Gaussian noise and progressively generates predictions. The encoder handles multimodal inputs, combining historical cyclone data with environmental field features containing both past and future information. At the core, the Physics-Inspired Decoder features the PIGA module, which uses cross-task attention to model the physical dependencies between task-specific features (trajectory, wind speed, and pressure), ensuring both physical consistency and accuracy in the forecasts.}
\label{fig:main}
\end{figure*}

\vspace{-0.4cm}
\section{Methodology}
\vspace{-0.35cm}
\label{sec:methodology}

We propose \textbf{Phys-Diff}, a physics-inspired Latent Diffusion Model \cite{rombach2022high} that integrates a Transformer-based encoder-decoder architecture \cite{vaswani2017attention} with physics-inspired inductive biases to generate accurate and reliable probabilistic forecasts.

\vspace{-0.4cm}
\subsection{Problem Formulation}
\vspace{-0.2cm}
\label{ssec:problem}

Given a historical sequence of TC observations over $M$ time steps, $\mathcal{H} = \{h_1, \ldots, h_M\}$, where each observation $h_i = (\mathbf{x}_i, v_i, p_i)$ consists of coordinates ($\mathbf{x}_i \in \mathbb{R}^2$), wind speed ($v_i \in \mathbb{R}$), and pressure ($p_i \in \mathbb{R}$). We are also provided with two sets of environmental fields: 1) historical fields from ERA5 reanalysis, $\mathcal{E}_{\text{hist}} = \{E_1, \ldots, E_M\}$, and 2) future forecast fields from the FengWu model \cite{chen2025fengwu}, $\mathcal{E}_{\text{fut}} = \{E'_{1}, \ldots, E'_{N}\}$, centered on a predicted future track. Each field $E \in \mathbb{R}^{C \times H \times W}$ contains $C=69$ variables at $H \times W = 80 \times 80$ spatial resolution. The objective is to predict the future TC sequence over $N$ time steps, $\mathcal{F} = \{f_1, \ldots, f_N\}$. In this paper, we set $M=4$, while $N$ varies according to the forecast horizon.

\vspace{-0.4cm}
\subsection{Dataset}
\vspace{-0.2cm}
\label{ssec:dataset}

We use the International Best Track Archive for Climate Stewardship (IBTrACS) dataset \cite{ibtracs2023} from 1980--2022 for TC trajectory and intensity ground truth. The environmental fields are sourced from ERA5 reanalysis data \cite{hersbach2020era5} (historical) and FengWu model forecasts (future). The data contains 69 variables (4 surface-level and 5 across 13 pressure levels) at a $0.25^\circ$ spatial resolution and a 6-hour temporal resolution. For model input, we crop a $10^\circ$ radius region centered on the TC from both ERA5 and FengWu fields. Future cropping centers for FengWu are determined using a TC tracking algorithm following \cite{bi2023accurate}. After cropping, each environmental input tensor has size $69 \times 80 \times 80$.

\vspace{-0.4cm}
\subsection{Phys-Diff Architecture}
\vspace{-0.2cm}

Fig.~\ref{fig:main} illustrates the Phys-Diff framework. As a physics-inspired Latent Diffusion Model built upon a Transformer encoder-decoder architecture, Phys-Diff operates in a learned latent space rather than directly on raw data, where diffusion and denoising processes achieve stronger representational capacity.

Formally, a convolutional encoder $\mathcal{E}$ maps the future TC sequence $x_0 \in \mathbb{R}^{N \times 4}$ into a latent representation $z_0 = \mathcal{E}(x_0)$, where $z_0 \in \mathbb{R}^{N \times D_{\text{embedding}}}$. The model then learns to reverse a diffusion process in this latent space. Finally, a decoder $\mathcal{D}$ maps the denoised latent representation $\hat{z}_0$ back to the data space to produce the forecast, $\hat{x}_0 = \mathcal{D}(\hat{z}_0)$.

Diffusion comprises a forward and a reverse process. The forward process gradually adds Gaussian noise to the latent representation $z_0$ over $T$ timesteps according to a fixed schedule $\bar{\alpha}_t$:
\begin{equation}
q(z_t | z_0) = \mathcal{N}(z_t; \sqrt{\bar{\alpha}_t} z_0, (1 - \bar{\alpha}_t) \mathbf{I})
\end{equation}

The reverse process, which is the core of our generative model, learns to remove this noise. Starting from pure Gaussian noise $z_T \sim \mathcal{N}(0, \mathbf{I})$, a denoising network $\epsilon_\theta$ iteratively predicts the added noise $\epsilon$ at each step $t$, conditioned on a context vector $c$. The model then generates a slightly cleaner latent variable $z_{t-1}$:
\begin{equation}
z_{t-1} = \frac{1}{\sqrt{\alpha_t}} \left( z_t - \frac{1 - \alpha_t}{\sqrt{1 - \bar{\alpha}_t}} \epsilon_\theta(z_t, t, c) \right) + \sigma_t \mathbf{w}
\end{equation}

where $\mathbf{w}$ is Gaussian noise (zero for $t=1$), $\alpha_t = 1-\beta_t$, and $\sigma_t$ is a fixed variance. The denoising network $\epsilon_\theta$ and the construction of the conditioning context $c$ are detailed below.

\vspace{-0.35cm}
\subsubsection{Conditional Denoising Network \texorpdfstring{$\epsilon_\theta$}{epsilon-theta}}
\vspace{-0.15cm}

The denoising network $\epsilon_\theta$ is a Transformer-based architecture composed of a conditional encoder and a physics-inspired decoder.

Conditional Encoder: The encoder's primary function is to fuse multimodal inputs into a unified context memory $c$ that guides the denoising process. Historical TC attributes $H$ are first encoded into a context token $H_{TC}$ using a Gated Recurrent Unit (GRU) \cite{cho2014learning}. In parallel, the historical and future environmental fields, $[E_{\text{hist}}, E_{\text{fut}}]$, are processed by a Swin Transformer \cite{liu2021swin} to produce a set of environmental feature tokens $T_{\text{env}}$. These tokens, along with a timestep embedding $t_{emb}$, are concatenated and processed by a standard Transformer encoder to allow for rich interaction between all conditioning variables. The final context $c$ is formulated as:
\begin{align}
c &= \text{TransformerEncoder}([\text{GRU}(H), \nonumber \\
&\quad \text{SwinTransformer}([E_{\text{hist}}, E_{\text{fut}}]), t_{emb}])
\end{align}

Physics-Inspired Decoder: The decoder predicts the noise $\epsilon$ from the noisy latent variable $z_t$, conditioned on the context $c$. Its architecture consists of a stack of decoder blocks. Each block first applies self-attention to its input sequence (derived from $z_t$) and then uses cross-attention to incorporate the conditioning information from the encoder's context memory $c$, producing a context-aware representation $X_{\text{cross}} \in \mathbb{R}^{N \times D_{\text{model}}}$. This representation is then refined by our core contribution, the PIGA module.

\vspace{-0.45cm}
\subsubsection{PIGA Module}
\vspace{-0.2cm}

Central to our decoder is the PIGA module, designed to explicitly model the physical interdependencies between TC attributes (trajectory, wind speed, and pressure). It is positioned after the cross-attention layer in each decoder block, ensuring that physical constraints inform the feature refinement process at multiple semantic levels.

The PIGA module operates on the context-aware feature $X_{\text{cross}}$ in four main steps:
\begin{enumerate}
\item \textbf{Decomposition:} $X_{\text{cross}}$ is projected into three task-specific streams: $f_{\text{traj}}$, $f_{\text{wind}}$, and $f_{\text{pres}}$, each with a dimension of $N \times D_{\text{sub}}$.
\item \textbf{Interaction:} Each stream attends to the other two using cross-task attention to capture physical relationships. For instance, the trajectory feature updates by attending to wind and pressure features.
\item \textbf{Gating:} An adaptive gate $g$ is computed to balance the influence of the original task-specific feature $f_{\text{orig}}$ and the new physics-informed feature from the attention step $A$.
\item \textbf{Fusion:} The updated, disentangled features are concatenated and fused via a $1 \times 1$ convolution to produce the final output $X_{\text{PIGA}}$, which is passed to the subsequent feed-forward network.
\end{enumerate}

The process for a single stream (e.g., trajectory) is formulated as:
\begin{align}
    f_{\text{traj}} &= \text{Proj}_{\text{traj}}(X_{\text{cross}}), \quad f_{\text{wind}} = \text{Proj}_{\text{wind}}(X_{\text{cross}}), \nonumber \\
    f_{\text{pres}} &= \text{Proj}_{\text{pres}}(X_{\text{cross}}) \\
    A_{\text{traj}} &= \text{Attention}(Q=f_{\text{traj}}, K,V=[f_{\text{wind}}, f_{\text{pres}}]) \\
    g_{\text{traj}} &= \sigma(\text{MLP}([f_{\text{traj}}, A_{\text{traj}}])) \\
    f'_{\text{traj}} &= (1 - g_{\text{traj}}) \odot f_{\text{traj}} + g_{\text{traj}} \odot A_{\text{traj}} \\
    X_{\text{PIGA}} &= \text{Conv}_{1 \times 1}(\text{Concat}(f'_{\text{traj}}, f'_{\text{wind}}, f'_{\text{pres}}))
\end{align}

This mechanism enables the model to learn disentangled yet physically-correlated representations, enhancing the physical consistency of the final forecast.

\vspace{-0.5cm}
\subsection{Training Objective}
\vspace{-0.25cm}

The model is trained with a composite objective combining the primary diffusion loss ($L_{\text{diffusion}}$) and an auxiliary reconstruction loss ($L_{\text{recon}}$). The diffusion loss minimizes the mean squared error between the true noise $\epsilon$ and the network prediction $\epsilon_\theta$. The reconstruction loss directly measures the error between the final prediction $\hat{x}_0$ and the ground truth $x_0$. The reconstruction loss is further decomposed as $L_{\text{recon}} = L_{\text{traj}} + L_{\text{wind}} + L_{\text{pres}}$, and task-specific gradient routing is enforced during backpropagation, where each component updates only its corresponding projection layer in the PIGA module, ensuring feature disentanglement.

To dynamically balance these components, we employ a learnable uncertainty-weighted scheme \cite{kendall2018multi}. The final training objective is formulated as:
\begin{equation}
L_{\text{total}} = \frac{1}{2\sigma_{\text{diff}}^2} L_{\text{diffusion}} + \frac{1}{2\sigma_{\text{recon}}^2} L_{\text{recon}} + \log(\sigma_{\text{diff}} \sigma_{\text{recon}})
\end{equation}

where $L_{\text{diffusion}} = \mathbb{E}_{t,z_0,\epsilon} \left[ \|\epsilon - \epsilon_\theta(z_t, t, c)\|^2 \right]$, and $\sigma_{\text{diff}}^2$ and $\sigma_{\text{recon}}^2$ are learnable parameters representing the uncertainty of each respective task.

\begin{table*}[!t]
\centering
\caption{Comprehensive TC forecast performance comparison across different basins and time horizons. Lower values indicate better performance. The GBRNN model predicts only trajectory, and MSCAR and VQLTI predict only intensity. $h$ represents hours, $s$ represents seconds, and inference time refers to a single sample.}
\label{tab:combined_performance}
\resizebox{\textwidth}{!}{%
\begin{tabular}{ll|cccc|cccc|cccc|cc}
\toprule
\multirow{2}{*}{\textbf{Basin}} & \multirow{2}{*}{\textbf{Methods}} & \multicolumn{4}{c|}{\textbf{Trajectory Error (km)}} & \multicolumn{4}{c|}{\textbf{Pressure Error (hPa)}} & \multicolumn{4}{c|}{\textbf{Wind Speed Error (m/s)}} & \multirow{2}{*}{\textbf{Model}} & \multirow{2}{*}{\textbf{Training/}} \\
\cmidrule(lr){3-6} \cmidrule(lr){7-10} \cmidrule(lr){11-14}
& & \textbf{6h} & \textbf{12h} & \textbf{18h} & \textbf{24h} & \textbf{6h} & \textbf{12h} & \textbf{18h} & \textbf{24h} & \textbf{6h} & \textbf{12h} & \textbf{18h} & \textbf{24h} & \textbf{size} & \textbf{Inference Time} \\
\midrule
\multirow{7}{*}{\textbf{Global}} & GRU(2014) & 53.30 & 101.99 & 190.01 & 298.62 & 4.10 & 5.50 & 7.20 & 9.10 & 3.20 & 4.10 & 5.30 & 6.80 & 322.88K & 3h/0.35s \\
& GBRNN (2019) & 42.91 & 65.34 & 103.77 & 147.12 & -- & -- & -- & -- & -- & -- & -- & -- & 1.50M & 11.2h/0.27s \\
& FengWu (2023) & 50.26 & 67.96 & 67.09 & 93.08 & 2.87 & 4.04 & 5.66 & 8.25 & 13.62 & 15.24 & 17.32 & 19.94 & 427M & --/0.58s \\
& ECMWF & \underline{41.50} & \underline{52.60} & \underline{65.80} & \underline{72.51} & 6.03 & 6.18 & 6.54 & 6.75 & -- & -- & -- & -- & -- & -- \\
& MSCAR(2024) & -- & -- & -- & -- & \textbf{2.28} & \underline{3.60} & \underline{4.66} & \underline{5.62} & \underline{1.64} & \underline{2.72} & 3.56 & \underline{4.30} & -- & -- \\
& VQLTI(2025) & -- & -- & -- & -- & 4.24 & 4.86 & 4.99 & 5.96 & 2.56 & 3.14 & \underline{2.77} & 4.45 & 2.90M & 10h/0.15s \\
\rowcolor{gray!20}
& \textbf{Phys-Diff (Ours)} & \textbf{15.26} & \textbf{17.25} & \textbf{41.84} & \textbf{54.35} & \underline{2.32} & \textbf{2.15} & \textbf{2.15} & \textbf{2.41} & \textbf{1.51} & \textbf{1.16} & \textbf{1.27} & \textbf{1.24} & 2.80M & 8h/0.74s \\
\midrule
\multirow{4}{*}{\textbf{WP}} & MMSTN(2022) & 28.10 & 60.25 & 98.10 & 142.30 & 1.75 & 2.90 & 4.05 & 4.85 & 1.99 & 2.15 & 2.18 & 2.60 & 4.80M & 10.8h/0.16s \\
& MGTCF(2023) & 23.90 & 44.10 & 68.20 & 94.50 & 1.80 & 2.15 & \underline{2.77} & 3.35 & 1.75 & 1.84 & \underline{1.58} & 1.90 & 3.60M & 12h/0.18s \\
& TC-Diffuser(2025) & \underline{21.35} & \underline{23.63} & \underline{49.95} & \underline{76.10} & \textbf{1.22} & \underline{2.04} & 2.15 & \underline{2.87} & \underline{1.19} & \underline{1.45} & 1.99 & \underline{1.73} & 9.50M & 6h/1.04s \\
\rowcolor{gray!20}
& \textbf{Phys-Diff (Ours)} & \textbf{15.19} & \textbf{17.32} & \textbf{42.07} & \textbf{54.45} & \underline{1.52} & \textbf{1.71} & \textbf{2.04} & \textbf{2.35} & \textbf{1.13} & \textbf{1.25} & \textbf{1.48} & \textbf{1.17} & 2.80M & 2.8h/0.74s \\
\bottomrule
\end{tabular}%
}
\end{table*}

\vspace{-0.6cm}
\begin{table}[!t]
\centering
\caption{Ablation study on key model components (lower is better). Since Phys-Diff is generative, we form an ensemble by sampling N=50 members from different Gaussian noise initializations; the ensemble mean improves accuracy over single-shot predictions.}
\label{tab:ablation}
\resizebox{1.0\columnwidth}{!}{%
\begin{tabular}{l|cccc|cccc|cccc}
\toprule
\multirow{2}{*}{\textbf{Settings}} & \multicolumn{4}{c|}{\textbf{Trajectory Error (km)}} & \multicolumn{4}{c|}{\textbf{Pressure Error (hPa)}} & \multicolumn{4}{c}{\textbf{Wind Speed Error (m/s)}} \\
\cmidrule(lr){2-5} \cmidrule(lr){6-9} \cmidrule(lr){10-13}
& \textbf{6h} & \textbf{24h} & \textbf{48h} & \textbf{120h} & \textbf{6h} & \textbf{24h} & \textbf{48h} & \textbf{120h} & \textbf{6h} & \textbf{24h} & \textbf{48h} & \textbf{120h} \\
\midrule
w/o PIGA & 18.15 & 65.40 & 105.70 & 211.30 & 2.85 & 3.12 & 3.95 & 5.51 & 1.93 & 1.74 & 2.56 & 3.42 \\
w/o FengWu & \underline{15.12} & 54.85 & 95.25 & 182.60 & 2.36 & 2.45 & 3.23 & 4.58 & 1.54 & \underline{1.23} & 2.15 & 2.88 \\
w/o both & 30.25 & 70.10 & 115.50 & 235.80 & 4.07 & 4.61 & 5.89 & 8.63 & 2.74 & 2.53 & 3.89 & 5.37 \\
\midrule
\rowcolor{gray!20}
\textbf{Phys-Diff} & 15.26 & \underline{54.35} & \underline{81.11} & \underline{151.80} & \underline{2.32} & \underline{2.41} & \underline{3.18} & \underline{4.42} & \underline{1.51} & 1.24 & \underline{2.08} & \underline{2.78} \\
\rowcolor{gray!20}
\textbf{Phys-Diff (Ensemble)} & \textbf{14.82} & \textbf{52.18} & \textbf{79.20} & \textbf{145.50} & \textbf{2.15} & \textbf{2.28} & \textbf{2.95} & \textbf{4.08} & \textbf{1.38} & \textbf{1.12} & \textbf{1.92} & \textbf{2.58} \\
\bottomrule
\end{tabular}%
}
\end{table}

\vspace{-0.1cm}
\section{EXPERIMENTS}
\vspace{-0.05cm}
\label{sec:experiments}

\vspace{-0.08cm}
\subsection{Experimental Setup}
\vspace{-0.08cm}

\vspace{-0.1cm}
\subsubsection{Dataset and Preprocessing}
\vspace{-0.1cm}
We use global TC data from 1980 to 2022. The ground truth for TC trajectory and intensity is sourced from the IBTrACS dataset. Environmental context is provided by 69 atmospheric variables from ERA5 reanalysis data and FengWu forecast fields, extracted within a $10^\circ$ radius of the TC center. The dataset is chronologically split into a training set (1980--2017), a validation set (2018), and a test set (2019--2022).

To ensure training stability, input variables are normalized using two methods. Trajectory coordinates $x_i$ are normalized relative to the initial point $x_{ref}$ to make the model invariant to the starting location ($x_{rel,i} = (x_i - x_{ref})/\sigma_{coord}$). TC intensity attributes (wind speed, pressure) and all environmental fields are independently normalized using their mean ($\mu_a$) and standard deviation ($\sigma_a$) from the training set ($a_{norm} = (a - \mu_a)/\sigma_a$).

\vspace{-0.4cm}
\subsubsection{Evaluation Metrics}
\vspace{-0.15cm}
We use Mean Absolute Error (MAE) as the primary metric, consistent with standard practice. Trajectory error (km) is the great-circle distance between predicted and ground truth coordinates using the Haversine formula. Pressure error (hPa) is the absolute difference in minimum sea-level pressure. Wind speed error (m/s) is the absolute difference in maximum sustained wind speed.

\vspace{-0.4cm}
\subsubsection{Implementation Details}
\vspace{-0.15cm}
Phys-Diff is implemented in PyTorch and trained on a single NVIDIA RTX 4090 GPU. We use the Adam optimizer with an initial learning rate of $1 \times 10^{-4}$ and a cosine annealing schedule. Training runs for 30 epochs with a batch size of 64. The training objective combines diffusion and reconstruction losses via uncertainty weighting.

\vspace{-0.43cm}
\subsection{Comparison with State-of-the-Art Methods}
\vspace{-0.18cm}

We benchmark Phys-Diff against a range of baseline models, including traditional DL methods (GRU \cite{cho2014learning}, GBRNN \cite{gao2019gbrnn}), recent models (MSCAR \cite{wang2024mscar}, VQLTI \cite{Wang_Liu_Chen_Han_Li_Bai_2025}, MMSTN \cite{mmstn2022}, MGTCF \cite{mgtcf2023}, TC-Diffuser \cite{zhang2025tc}), large-scale models (FengWu \cite{chen2025fengwu}), and operational NWP systems (ECMWF). We also evaluate on the Western North Pacific (WP) basin, the world's most active TC region with highly complex weather systems. As shown in Table~\ref{tab:combined_performance}, Phys-Diff consistently
outperforms all baseline models in trajectory forecasting across both
Global and WP basins. For the Global basin at the 24-hour forecast horizon, Phys-Diff reduces the trajectory error by 25.0\% compared to the operational ECMWF model. In the highly active WP
basin, this advantage is even more significant, with our model reducing the 24-hour error by 28.5\% compared to the second best model
TC-Diffuser. For intensity forecasting, Phys-Diff achieves superior performance at longer forecast horizons, attaining the lowest error for both pressure and wind speed at 24 hours across both basins.
On the Global dataset, Phys-Diff reduces the 24-hour pressure forecast error by 57.1\% compared to MSCAR and the wind speed forecast error by 71.2\% compared to MSCAR. In the WP basin, our model lowers the 24-hour pressure error by 18.1\% and the wind speed error by 32.4\% compared to TC-Diffuser. This robust and
consistent performance demonstrates the effectiveness of Phys-Diff in accurately modeling TC dynamics and
mitigating error accumulation over extended periods.

\vspace{-0.38cm}
\subsection{Ablation Study}
\vspace{-0.13cm}

We conduct an ablation study on the PIGA module and FengWu forecast data. The results are presented in Table~\ref{tab:ablation}.

\vspace{-0.35cm}
\subsubsection{Quantitative Analysis}
\vspace{-0.1cm}
The results demonstrate the critical role of each component. Removing the PIGA module (`w/o PIGA') substantially degrades performance, with the 24h trajectory error increasing by 20.3\%, confirming its role in enforcing physical consistency. With only historical data (`w/o FengWu'), the model remains robust, but incorporating future fields from FengWu improves long-term accuracy. Additionally, we leverage the generative nature of Phys-Diff to form an ensemble by sampling N=50 members from different Gaussian noise initializations. As shown in Table~\ref{tab:ablation}, the ensemble mean consistently improves accuracy over single-shot predictions, demonstrating the model's capability for uncertainty quantification.

\vspace{-0.35cm}
\subsubsection{Qualitative Analysis}
\vspace{-0.1cm}

\begin{wrapfigure}{l}{0.25\textwidth}
\centering
\includegraphics[width=0.23\textwidth]{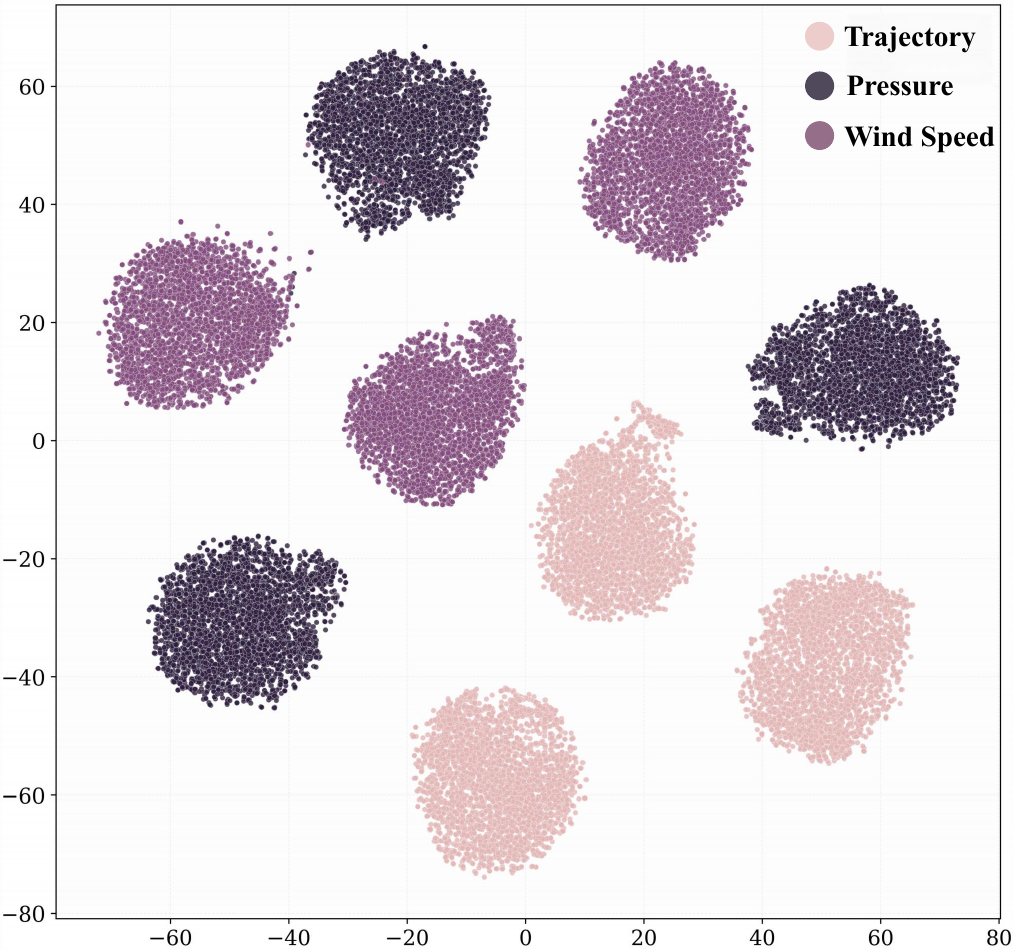}
\caption{t-SNE visualization of learned task-specific features by the PIGA module.}
\label{fig:tsne}
\end{wrapfigure}

\begin{figure}[t]
\centering
\includegraphics[width=0.48\textwidth]{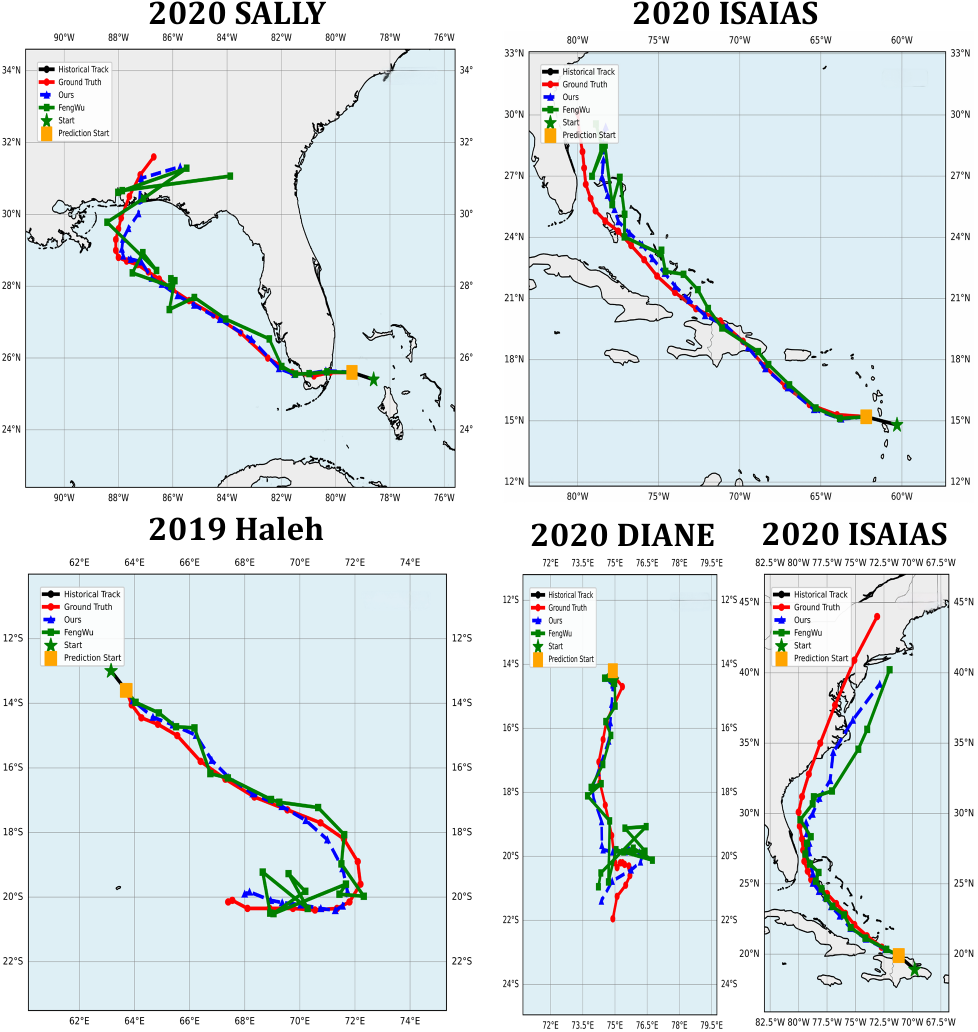}
\vspace{-12pt}
\caption{Visualization of our trajectory forecasting results and comparison with FengWu. The five small plots visualize the ground truth (red solid line), our trajectory forecasting results (blue dashed line), and FengWu's trajectory forecasting results (green solid line), with representative trajectories selected.}
\label{fig:track}
\end{figure}

Fig.~\ref{fig:track} visualizes representative trajectory predictions compared with FengWu across challenging scenarios (linear variation, sudden turning, spiral variation, and land interaction). Our method better captures complex trajectory patterns.

We also visualize PIGA's learned features using t-SNE \cite{maaten2008visualizing} (Fig.~\ref{fig:tsne}). Three distinct clusters emerge: trajectory (pink), pressure (black), and wind speed (purple). Pressure and wind speed clusters are spatially closer with overlap, consistent with their physical coupling. The trajectory cluster remains distant, reflecting its distinct positional nature.

\vspace{-0.4cm}
\section{CONCLUSION}
\vspace{-0.35cm}

In this paper, we presented Phys-Diff, a physics-inspired latent diffusion model for tropical cyclone forecasting. Our model explicitly captures the physical interdependencies between trajectory, pressure, and wind speed through the PIGA module, which learns disentangled feature representations via cross-task attention. Experiments show Phys-Diff achieves state-of-the-art performance on global and regional datasets, reducing 24-hour forecast errors by 41.6\% for trajectory, 57.1\% for pressure, and 71.2\% for wind speed compared to existing deep learning methods.

\vspace{-0.4cm}
\section{ACKNOWLEDGMENT}
\vspace{-0.35cm}

This research is supported by Smart-Grid National Science and Technology Major Project (Grant No. 2025ZD0805500).

\vspace{-8pt}
\bibliographystyle{config/IEEEbib}
\bibliography{bib/strings,bib/refs}

\end{document}